\documentclass[runningheads]{llncs}

\usepackage[T1]{fontenc}
\usepackage{graphicx}
\usepackage{booktabs}
\usepackage{amsmath}
\usepackage{multirow}
\usepackage{tikz}
\usetikzlibrary{positioning,arrows.meta,calc}
\usepackage{pifont}
\usepackage[hidelinks]{hyperref}
\definecolor{pacegreen}{RGB}{14,122,99}
\definecolor{pacered}{RGB}{176,72,58}
\definecolor{pacegray}{RGB}{120,128,124}

\newcommand{\impro}[1]{{\hspace{0.05cm}\textcolor{pacegreen}{\textbf{(+#1)}}}}
\newcommand{\improx}{${}_{\hspace{0.05cm}\hphantom{\textbf{(+0.0)}}}$}
\newcommand{\detro}[1]{{\hspace{0.05cm}\textcolor{pacered}{\textbf{($-$#1)}}}}

\begin{document}

\title{PACE: Plan, Attend, Check, and Emit for Training-Free Response Timing in Streaming Video Understanding}

\titlerunning{PACE}

\author{Siqi Guo\inst{1} \and
Yuxia Chen\inst{2} \and
Wenbo Zhu\inst{3} \and
Yangguang Ji\inst{4} \and
Zhenxiang Jiang\inst{5} \and
Teng Zhong\inst{6} \and
Yu Zhang\inst{7}\thanks{Corresponding author.}}
\authorrunning{S. Guo et al.}

\institute{Nanyang Technological University \and
University of Electronic Science and Technology of China \and
University of California, Berkeley \and
Nanyang Technological University \and
National University of Singapore \and
Nanjing Normal University \and
Beijing Freedo Technology}
\maketitle

\begin{abstract}
Streaming video understanding requires a model to decide not only what to answer but also when to answer. Evidence arrives progressively, and recent benchmarks identify premature responses, issued before the queried event has concluded, as a recurring failure mode. Existing approaches often learn response timing through additional training, while training-free alternatives typically rely on scalar gating signals that do not explicitly represent when an answer becomes final. We organize chunk-wise streaming errors into three patterns: sampling variation can make an answer unstable, a stable answer may lack support in the observed video, or an answer may be supported by the current evidence while the event remains unfinished. We propose PACE (\textbf{P}lan, \textbf{A}ttend, \textbf{C}heck, and \textbf{E}mit), a training-free framework designed around these three failure modes. Given a question, PACE drafts a plan specifying what to observe and what will signal completion. As the stream unfolds, it invokes an off-the-shelf video LLM at informative moments, evaluates each draft for stability, grounding in its cited frames, and completion, and records verified observations in an event ledger. Experiments on ProactiveVideoQA and OVO-Bench Forward show that PACE improves over the official protocols by up to 5.9 average PAUC points and 6.2 average accuracy points across proprietary and open-source backbones, yielding state-of-the-art performance. 
\end{abstract}

\begin{figure}[t]
  \centering
  \includegraphics[width=\linewidth]{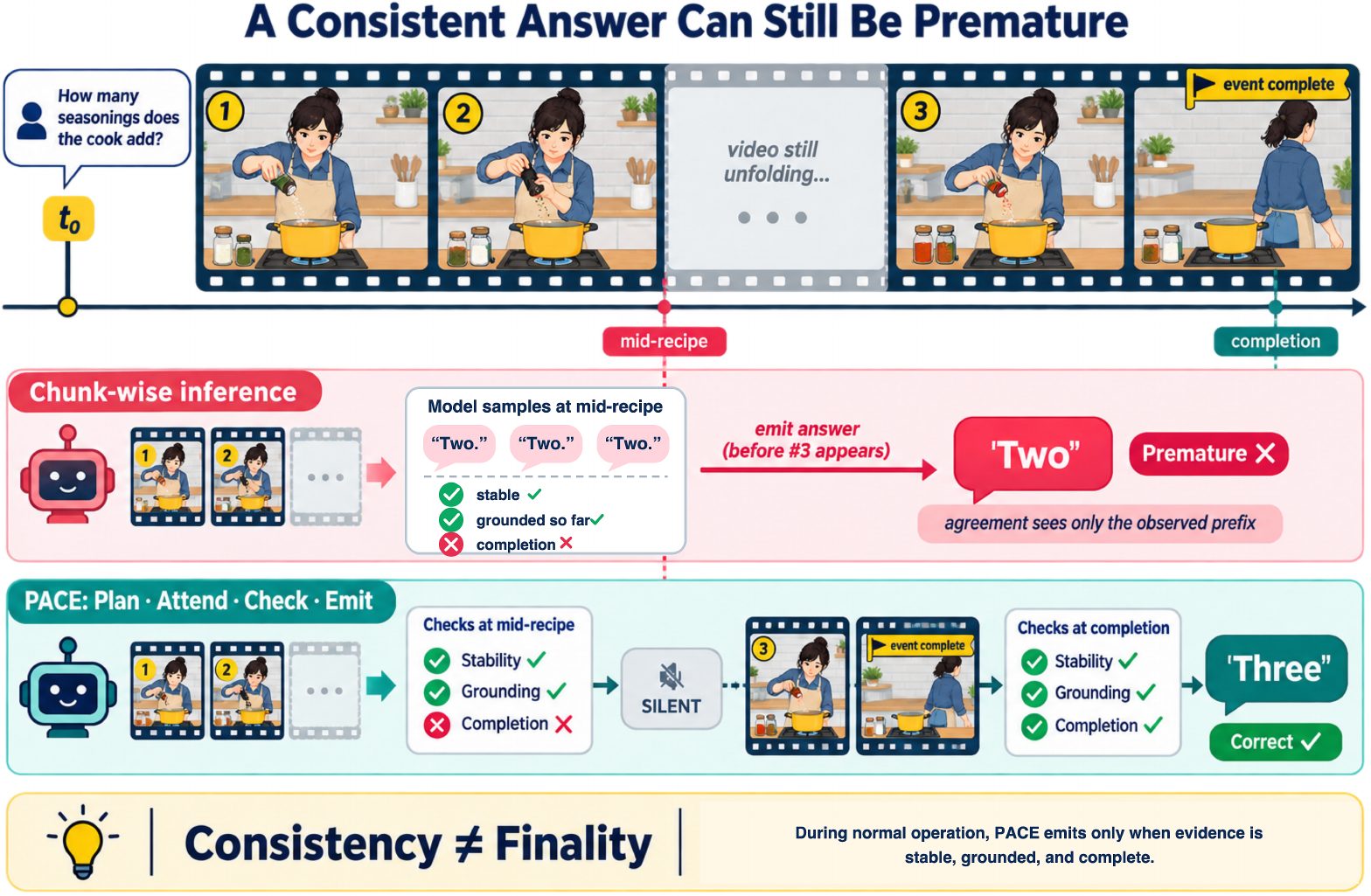}
  \caption{Response timing on a video stream: evidence for a question posed
  at $t_0$ arrives over time. A chunk-wise baseline answers at every chunk;
  its first two replies are premature, and it cannot know that its third is
  final. PACE holds the draft until its watch plan's completion
  signal fires, then answers once, correctly.}
  \label{fig:teaser}
\end{figure}

\section{Introduction} \label{sec:intro}

Video large language models are increasingly being considered for live-streaming settings, including narrating an egocentric camera, monitoring a surveillance feed, and answering questions during playback~\cite{videollmonline,dispider,livestar,videochatonline,flashvstream}. In these settings, video arrives continuously and interaction takes place while the content is still unfolding. Unlike offline video question answering, where the full video is available before inference, streaming requires a model to decide not only \emph{what} to answer but also \emph{when} to answer and when to remain silent, as relevant evidence may arrive seconds or minutes after a question is asked (Fig.~\ref{fig:teaser}).

Most prior work learns response timing through silence-token objectives, scoring heads, or dedicated decision modules~\cite{videollmonline,mmduet,mmduet2,dispider,livestar,streambridge}, while existing training-free alternatives gate responses using scalar signals such as an embedding delta or a token-drop ratio~\cite{ipibench,timechatonline}. These approaches have complementary limitations. Learned timing components are difficult to transfer directly to proprietary black-box models, even though those models perform strongly under naive chunk-wise inference on ProactiveVideoQA~\cite{proactivevideoqa}. Scalar gating signals, meanwhile, do not explicitly represent whether the evidence required by a particular question is complete.

Across recent benchmarks, premature answering emerges as a prominent error. OVO-Bench asks a model to withhold its response until enough future evidence has arrived and finds this ability to be the weakest of the three it evaluates~\cite{ovobench}; StreamingBench and OmniMMI report a related offline-to-streaming gap~\cite{streamingbench,omnimmi}; and 92.7\% of Gemini~3~Pro's incorrect proactive triggers on IPIBench occur before the queried event has finished~\cite{ipibench}. These results suggest that response timing, rather than visual perception alone, is often the immediate source of error. Self-consistency~\cite{selfconsistency} does not directly address this issue: for the question in Fig.~\ref{fig:teaser}, when asked mid-recipe, a model may repeatedly answer ``two,'' which is consistent with the evidence observed so far but incorrect as a final answer because a third seasoning has yet to appear. Finality therefore depends on both the semantics of the question and the progress of the event; agreement among successive answers alone does not establish it.

We organize chunk-wise streaming errors into three recurring failure modes: decoding noise, stable answers that lack visual support, and supported but premature answers. Based on this taxonomy, we propose PACE (\textbf{P}lan, \textbf{A}ttend, \textbf{C}heck, and \textbf{E}mit), a training-free framework that coordinates calls to an off-the-shelf video LLM and applies a corresponding check to each failure mode. Given the question, \emph{Plan} drafts a watch plan that specifies what to observe and which visual signal will indicate completion; the plan can subsequently be revised under constrained rules. \emph{Attend} schedules backbone calls at informative moments in the stream. \emph{Check} evaluates each draft for stability, grounding, and completion: stability voting reduces sensitivity to decoding noise, re-derivation from cited frames targets answers without visual support, and the completion check tests whether a consistent answer is still premature. \emph{Emit} records replies, supporting evidence, and plan revisions in an event ledger, reducing duplicate outputs and carrying prior evidence forward.

Experiments on ProactiveVideoQA and OVO-Bench Forward show that PACE improves over the official protocols by up to 5.9 average PAUC points and 6.2 average accuracy points across proprietary and open-source backbones, yielding state-of-the-art performance.

Our contributions are as follows:
\begin{itemize}
\item We organize the premature responses of streaming video LLMs into a three-way taxonomy, and identify consistent prematureness as a stream-specific failure that no inspection of the answer itself can detect.
\item We propose PACE, a training-free and model-agnostic framework whose checks match the taxonomy one-to-one, built around a self-revising watch plan and an auditable event ledger.
\item PACE achieves state-of-the-art results on ProactiveVideoQA and OVO-Bench Forward across proprietary and open-source backbones.
\end{itemize}

\section{Related Work} \label{sec:related}

\subsubsection{Video Large Language Models.}

Modern video LLMs such as the Qwen-VL series and LLaVA-Video combine a vision encoder with a strong language model and are tuned on large video instruction corpora~\cite{qwen25vl,zhu2025dynamic,qwen3vl,zhang2024llava}. Their capabilities extend from temporal grounding~\cite{wu2025number} and user-generated-video repurposing~\cite{wu2025video} to intelligent image editing~\cite{wu2026kris}. Parallel visual research preserves edge and frequency detail in remote sensing and MRI~\cite{chen2024hi,fang2026chasm,chen2026hifi}, or organizes representations through hierarchical semantic alignment~\cite{zhu2026hierarchical}. Agentic methods further plan frame selection and tool use over a complete video~\cite{videoagent}. Most of these systems assume that the full video is available before generation, while several high-performing models are accessible only through black-box APIs. PACE instead considers frames that arrive causally in a stream.

\subsubsection{Streaming Video Understanding.}
Recent benchmarks report a substantial gap between offline and streaming video understanding. Models that perform well with complete videos often degrade when frames arrive incrementally, and withholding a response until future evidence becomes available remains particularly challenging~\cite{streamingbench,omnimmi,proactivevideoqa,ovobench}. One line of work incorporates streaming behavior through interleaved silence and response, streaming memory, system-level designs, or response-timing heads~\cite{videollmonline,flashvstream,videochatonline,ixc25ol,dispider,mmduet,mmduet2,streambridge,vispeak}.


Training-free methods reduce inference cost through KV-cache retrieval and token dropping~\cite{rekv,timechatonline}, or schedule perception around anticipated evidence~\cite{m3agent,streamagent}. More broadly, adaptation has been pursued through feature regularization and label-free prompting~\cite{zhu2024boosting,zhu2024enhancing}, selective vision-language subspaces~\cite{zhu2024selective}, cross-modal alignment~\cite{zhu2025enhancing}, and test-time prompt updates~\cite{zhu2026robustifying}. Analyses of in-context learning and SFT generalization further motivate flexible inference-time control~\cite{wu2024glance,wu2026generalization}. Methods that explicitly model response release commonly use a decoding-time score~\cite{livestar}, a token-drop ratio~\cite{timechatonline}, or similarity between watch proposals and recent frames~\cite{ipibench}. These signals do not explicitly encode question-dependent completion conditions. Online action detection estimates event boundaries~\cite{oad}, but not whether the evidence needed for a particular answer is complete.


PACE is also related to test-time verification methods based on repeated sampling, consistency, or visual reinforcement~\cite{selfconsistency,zhu2026lookcarefully,selfcheckgpt,cove}. Evidence-based approaches mitigate hallucinations or abstain when an answer is unreliable~\cite{reliablevqa,zhu2026mitigating,selectiveqa}, while complementary interventions target safety alignment~\cite{zhu2026guardalign}, jailbreak defense~\cite{zhu2026principled}, or removal of unwanted concepts~\cite{wu2025unlearning}. These methods operate under a fixed context. In a stream, however, a currently supported answer need not be final~\cite{ovobench,ipibench}. PACE therefore adds constrained watch-plan revision, an explicit completion check, and an event ledger for verified observations.

\section{Method} \label{sec:method}

\begin{figure*}[t]
\centering
\includegraphics[width=\textwidth]{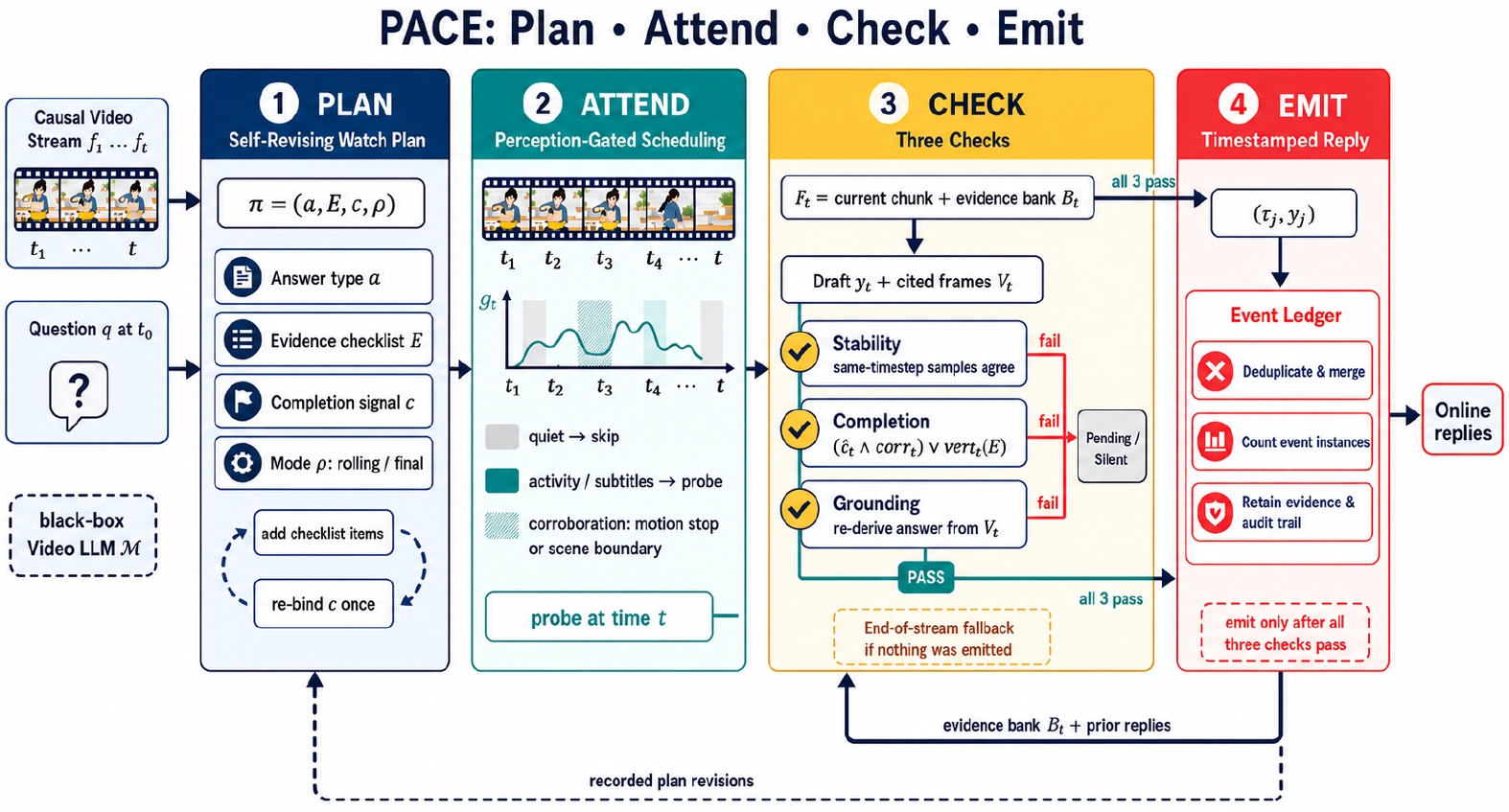}
\caption{PACE plans what to observe, schedules activity-triggered probes, checks each draft, and records verified events and evidence for subsequent probes.}
\label{fig:framework}
\end{figure*}

Sect.~\ref{sec:setup} formalizes the problem and the error taxonomy; Sects.~\ref{sec:plan} to \ref{sec:emit} then present the four stages of PACE shown in Fig.~\ref{fig:framework}.

\subsection{Preliminaries} \label{sec:setup}

A video stream is a frame sequence $\{f_1, f_2, \dots\}$ revealed causally: at time $t$ the system observes only the prefix $f_{\le t}$. The backbone is a video LLM accessed as a black box, a function $\mathcal{M}$ that maps a frame set $F$ and a text prompt $p$ to a text output $\mathcal{M}(F, p)$, decoded greedily or sampled; PACE interacts with the backbone only through such calls. A question $q$ arrives at time $t_0$, and the system outputs timestamped replies $\{(\tau_j, y_j)\}$ online: at any step it may stay silent or emit a reply, and emitted replies cannot be retracted.

In our analysis of chunk-wise baselines, incorrect replies follow three recurring patterns: same-timestep resamples disagree because of decoding noise; a stable reply lacks support in the video observed so far; or a stable, supported reply concerns an event that has not yet concluded. These underlying faults can overlap in a single reply. For analysis, we therefore assign labels with a fixed decision tree, disagreement first, followed by support and completion, and give each incorrect reply the first applicable label.

\subsection{Plan: a Self-Revising Watch Plan} \label{sec:plan}

When $q$ arrives, one text-only call $\mathcal{M}(\emptyset,q)$ produces a watch plan $\pi=(a,E,c,\rho)$: answer type $a$, evidence checklist $E$, completion signal $c$, and reporting mode $\rho\in\{\text{rolling},\text{final}\}$. State questions admit a current answer, cumulative questions grow until $c$, and reveal questions become answerable only after the relevant evidence appears. Monitor instructions reuse these types and produce one reply per event instance. The question wording determines $\rho$: definite totals and outcomes use final mode, while descriptive requests use rolling mode. The type and mode remain fixed throughout the stream.

PACE permits two plan revisions. Checklist items may be added to $E$ at any probe. The completion signal may be re-bound once, when the probe reports a setting mismatch and the scheduler detects activity in the preceding $k$ chunks. The replacement must be observable by the scheduler. Each revision and its reason are recorded in the ledger; all other plan components remain unchanged.

\subsection{Attend: Perception-Gated Scheduling} \label{sec:attend}

The stream advances in fixed chunks, and a lightweight activity signal $g_t$, the mean absolute difference of consecutive frames, determines when to invoke the backbone. Quiet stretches are skipped up to a bounded maximum silence, while increased activity raises the probe density. Two signal sets serve different purposes. The probe-trigger set combines pixel and subtitle changes, with subtitles taken from the benchmark's own track, and uses permissive thresholds to retain probes for dialogue-driven reveals. The corroboration set comprises motion cessation or a scene boundary after sustained activity and uses stricter thresholds to reduce completion triggers from isolated cuts in fast-edited content. For monitor questions, an onset is defined as activity rising above the probe threshold after a quiet stretch. These signals control probe timing and provide corroboration, but emission remains subject to the checks in the next stage.

\subsection{Check: the Three Checks} \label{sec:check}

At probe time $t$, PACE makes a greedy call $\mathcal{M}(F_t,p_t)$, where $F_t$ contains the current chunk and evidence bank $B_t$, and $p_t$ contains $q$, plan $\pi$, and prior replies. The call returns a draft $y_t$, cited frames $V_t\subseteq F_t$, checklist progress $\hat{E}_t$, completion flag $\hat{c}_t$, setting match, and, for cumulative or monitor questions, newly observed event instances with cited frames. Abstentions and drafts with no new information produce no output. Otherwise, the draft must pass the following three checks during normal operation.

\paragraph{Stability.} A draft is new if it differs from both the last emitted reply and the pending draft under a type-aware comparator; the first draft is always new. For a new draft, PACE draws $n-1$ additional samples at the same timestep and compares all samples with the greedy draft. Counts use numeric equality, multiple-choice answers use option identity, and free text uses token-level F1 above $\theta_2$. The draft passes when at least a fraction $\theta_1$ agree. A draft that passes stability but fails a later check becomes pending. If the same pending draft recurs at the next probe, it passes stability without resampling; a draft that previously failed stability is sampled again.

\paragraph{Completion.} State questions and rolling reports, including rolling counts, pass by definition. A final reply to a cumulative or reveal question passes if and only if
\begin{equation}
\bigl(\hat{c}_t \wedge \mathrm{corr}_t\bigr) \;\vee\; \mathrm{ver}_t(E),
\label{eq:completion}
\end{equation}
where $\mathrm{corr}_t$ indicates a candidate completion event since the previous probe. The term $\mathrm{ver}_t(E)$ holds when every checklist item $e\in E$ is confirmed by a context-isolated call $\mathcal{M}(F_t,e)$ that does not receive $y_t$ or $\hat{E}_t$. For cumulative and reveal questions, $E$ includes a completion item. Monitor replies use the same rule per event instance: an emitted and corroborated instance closes, and the next corroborated onset opens a new one. Completion prompts ask whether the relevant activity has visibly ended, rather than whether the model knows the answer.

\paragraph{Grounding.} A context-isolated call $\mathcal{M}(V_t,q)$ re-derives the answer from the cited frames and question. The draft passes if the result matches $y_t$ under the same comparator with threshold $\theta_2$; a draft with no cited frames fails. For cumulative questions, grounding is applied to each ledger increment rather than to the final sum. Using the same backbone for drafting and verification may preserve a shared misperception.

\paragraph{End-of-stream fallback.} If the stream ends before any reply is emitted, PACE releases the pending draft with the highest vote share. If no draft passed stability, it releases the last greedy draft. This fallback can therefore release a draft that has not passed all three checks.

\subsection{Emit: the Event Ledger} \label{sec:emit}

Every emitted reply $(\tau_j, y_j)$, event observation, and plan revision enters the ledger at the end of its probe. The ledger has three functions. First, it reduces duplication: drafts sufficiently similar to emitted content are suppressed, and an observation that overlaps an existing entry in time and cited frames is merged rather than added. Second, it supports counting: each observation that passes the stability and grounding checks becomes an entry whether or not a reply is emitted, and cumulative totals are read from the ledger rather than re-counted from pixels at every probe, a known difficulty in streaming counting~\cite{ovobench}. Third, it retains evidence for subsequent probes and audit: cited frames from replies and entries form the evidence bank $B_t$, which uses a fixed capacity with oldest-first eviction, and each plan revision is stored with its reason.

\section{Experiments} \label{sec:exp} 

\subsection{Setup} \label{sec:exp-setup} 

\paragraph{Benchmarks and protocol.} We evaluate on ProactiveVideoQA~\cite{proactivevideoqa} and OVO-Bench Forward Active Responding~\cite{ovobench}. ProactiveVideoQA contains 1{,}377 videos and 3{,}510 reply turns from web videos, egocentric videos, TV series, and anomaly detection videos. We use its official PAUC metric with $\omega{=}0.5$ and the official GPT-4.1 judge. OVO-Bench evaluates repetition counting, step recognition, and clues-reveal responding with per-probe accuracy. We follow the official temporal settings for both benchmarks. ProactiveVideoQA uses 2\,s chunks at 2\,fps for web videos and 5\,s chunks at 1\,fps for the other domains. OVO-Bench uses fixed probe timestamps. At each OVO-Bench probe, PACE reports the rolling ledger count for repetition counting, the current state for step recognition, and yes for clues-reveal responding when all checks pass. Only the clues-reveal task uses the completion check, while model-drafted question types are evaluated only on ProactiveVideoQA. 

\paragraph{Models, baselines, and implementation.} On ProactiveVideoQA, we evaluate PACE with Qwen2.5-VL-7B~\cite{qwen25vl}, GPT-4.1, and GPT-4.1-mini. On OVO-Bench, we use Qwen2-VL-7B~\cite{qwen2vl}, GPT-4o~\cite{gpt4o}, and Gemini 1.5 Pro~\cite{gemini15}. On each benchmark the backbones are chosen among models with officially reported baselines, so every unwrapped number has a published reference. The default setting uses $n{=}3$, $\theta_1{=}2/3$, $\theta_2{=}0.6$, a re-binding window of $k{=}3$ chunks, vote temperature 0.7, and greedy temperature 0. Maximum probe silence is 20\,s and is reduced to 10\,s for anomaly detection videos. Other thresholds, prompts, frame budgets, and evidence-bank limits are provided with the code. We run stochastic configurations with three seeds and report the mean. The main tables compare PACE with published systems and pair each PACE result with its unmodified backbone; all matched comparisons use the same chunking and frame budget. \begin{table}[!t]
\centering
\small
\caption{PAUC on ProactiveVideoQA~\cite{proactivevideoqa} with $\omega{=}0.5$. Published results are taken from the corresponding papers. The bottom block compares each backbone with and without PACE.}
\label{tab:pauc}
\begin{tabular}{@{}lccccc@{}}
\toprule
Model or method & WEB & EGO & TV & VAD & Avg. \\
\midrule
Human~\cite{proactivevideoqa} & 38.6\improx & 38.2\improx & 47.0\improx & 53.6\improx & 44.4\improx \\
\midrule
LLaVA-OneVision-7B~\cite{llavaonevision} & 55.0\improx & 61.6\improx & 45.1\improx & 25.6\improx & 46.8\improx \\
VideoLLM-online~\cite{videollmonline} & 25.9\improx & 25.0\improx & 18.3\improx & 25.0\improx & 23.6\improx \\
MMDuet~\cite{mmduet} & 38.9\improx & 46.0\improx & 21.1\improx & 27.4\improx & 33.4\improx \\
MMDuet2-RL~\cite{mmduet2} & 53.3\improx & 33.6\improx & 43.4\improx & 28.9\improx & 39.8\improx \\
Em-Garde~\cite{emgarde} & 44.3\improx & 52.3\improx & --\improx & 27.4\improx & --\improx \\
VideoChat3-4B~\cite{videochat3} & 28.4\improx & 28.1\improx & 34.7\improx & 25.1\improx & 29.1\improx \\
MOSS-VL-Realtime~\cite{mossvl} & 55.4\improx & 47.8\improx & 50.2\improx & 35.3\improx & 47.2\improx \\
\midrule
\multicolumn{6}{c}{PACE with Different Backbones} \\
GPT-4.1~\cite{proactivevideoqa} & 51.7\improx & 58.8\improx & 56.8\improx & 46.2\improx & 53.4\improx \\
\quad \textbf{+ PACE} & $59.2_\impro{7.5}$ & $64.7_\impro{5.9}$ & $62.1_\impro{5.3}$ & $51.1_\impro{4.9}$ & $59.3_\impro{5.9}$ \\
GPT-4.1-mini~\cite{proactivevideoqa} & 47.8\improx & 65.8\improx & 59.4\improx & 47.7\improx & 55.2\improx \\
\quad \textbf{+ PACE} & $54.1_\impro{6.3}$ & $70.2_\impro{4.4}$ & $62.8_\impro{3.4}$ & $56.9_\impro{9.2}$ & $61.0_\impro{5.8}$ \\
Qwen2.5-VL-7B~\cite{qwen25vl} & 52.7\improx & 46.5\improx & 36.5\improx & 29.3\improx & 41.3\improx \\
\quad \textbf{+ PACE} & $59.9_\impro{7.2}$ & $54.8_\impro{8.3}$ & $40.1_\impro{3.6}$ & $33.7_\impro{4.4}$ & $47.1_\impro{5.8}$ \\
\bottomrule
\end{tabular}
\end{table}

\subsection{Main Results} \label{sec:exp-main} \textbf{PACE lifts every backbone on ProactiveVideoQA.} Table~\ref{tab:pauc} shows consistent gains: GPT-4.1 improves from 53.4 to 59.3, GPT-4.1-mini from 55.2 to 61.0, and Qwen2.5-VL-7B from 41.3 to 47.1. The average gain is nearly constant, 5.8 to 5.9 points, although the three backbones differ in scale, family, and access, and the weakest of them benefits as much as the strongest. The wrapped GPT-4.1-mini sets the best average reported on the benchmark, 5.8 points above the previous best, its own backbone under the official chunk-wise protocol, and 13.8 points above the strongest purpose-built streaming system. The human reference row trails several systems, an artifact of live annotation latency under a metric that penalizes delayed replies. 

\textbf{Where the gains land differs by backbone.} For GPT-4.1 and Qwen2.5-VL-7B the largest improvements come on WEB and EGO, up to 7.5 and 8.3 points, where frequent activity keeps rolling answers current and the ledger suppresses repetition over long horizons. GPT-4.1-mini instead gains most on VAD, 9.2 points: anomaly clips reward a system that stays silent until the anomaly has run its course, and the completion machinery converts that patience into score. TV shows the smallest gains for GPT-4.1-mini and Qwen2.5-VL-7B, 3.4 and 3.6 points, and is among the smallest for GPT-4.1, suggesting that dialogue-driven reveals remain the hardest timing problem. \begin{table}[!t]
\centering
\small
\caption{Accuracy on OVO-Bench Forward Active Responding~\cite{ovobench}. Published results are taken from the corresponding papers. The bottom block compares each backbone with and without PACE.}
\label{tab:ovo}
\begin{tabular}{@{}lcccc@{}}
\toprule
Model or method & REC & SSR & CRR & Avg. \\
\midrule
Human~\cite{ovobench} & 95.5\improx & 89.7\improx & 93.6\improx & 92.9\improx \\
\midrule
LLaVA-Video-7B~\cite{zhang2024llava} & 34.1\improx & 70.0\improx & 60.4\improx & 54.8\improx \\
Qwen2-VL-72B~\cite{qwen2vl} & 38.8\improx & 64.1\improx & 45.0\improx & 49.3\improx \\
Flash-VStream-7B~\cite{flashvstream} & 8.0\improx & 67.3\improx & 60.0\improx & 45.1\improx \\
Dispider-7B~\cite{dispider} & 18.1\improx & 37.4\improx & 48.8\improx & 34.7\improx \\
Em-Garde~\cite{emgarde} & --\improx & --\improx & --\improx & 31.0\improx \\
MOSS-VL-Realtime~\cite{mossvl} & --\improx & --\improx & --\improx & 62.1\improx \\
\midrule
\multicolumn{5}{c}{PACE with Different Backbones} \\
Gemini 1.5 Pro~\cite{gemini15} & 35.5\improx & 74.2\improx & 61.7\improx & 57.2\improx \\
\quad \textbf{+ PACE} & $45.1_\impro{9.6}$ & $79.9_\impro{5.7}$ & $65.2_\impro{3.5}$ & $63.4_\impro{6.2}$ \\
GPT-4o~\cite{gpt4o} & 27.6\improx & 73.2\improx & 59.4\improx & 53.4\improx \\
\quad \textbf{+ PACE} & $35.2_\impro{7.6}$ & $74.1_\impro{0.9}$ & $63.8_\impro{4.4}$ & $57.7_\impro{4.3}$ \\
Qwen2-VL-7B~\cite{qwen2vl} & 31.7\improx & 65.8\improx & 48.8\improx & 48.7\improx \\
\quad \textbf{+ PACE} & $33.9_\impro{2.2}$ & $70.5_\impro{4.7}$ & $50.2_\impro{1.4}$ & $51.5_\impro{2.8}$ \\
\bottomrule
\end{tabular}
\end{table}

\textbf{OVO-Bench Forward shows the same picture.} In Table~\ref{tab:ovo}, all three backbones improve, Gemini 1.5 Pro from 57.2 to 63.4, GPT-4o from 53.4 to 57.7, and Qwen2-VL-7B from 48.7 to 51.5, and the wrapped Gemini passes the strongest published system, MOSS-VL-Realtime at 62.1. For the two proprietary backbones the largest gains fall on repetition counting, 9.6 points for Gemini and 7.6 for GPT-4o, the subtask that probes cross-time memory and the one the ledger is built for. The weaker Qwen2-VL-7B instead gains most on step recognition, 4.7 points, where the stability vote filters its noisier drafts. Clues-reveal responding improves on every backbone, by 1.4 to 4.4 points. The gap to human performance remains wide, 63.4 against 92.9, and most of it sits on repetition counting. 

\textbf{Summary.} Three facts hold across both benchmarks. PACE improves all six backbones, by 2.8 to 6.2 average points, under one configuration shared across backbones and benchmarks, apart from the domain-level silence bound stated in Sect.~\ref{sec:exp-setup}. The wrapped systems set the best reported averages on both benchmarks. And the gains concentrate where each mechanism is designed to pay: counting subtasks for the ledger, event-bounded domains for the completion machinery, and noisy drafts for the stability vote. The strongest baseline also remains the strongest wrapped system, so PACE adds response timing on top of perception quality rather than trading one for the other. 

\subsection{Visualization} \label{sec:exp-vis} 
\begin{figure}[!t]
\centering
\small
\begin{minipage}[t]{0.235\textwidth}\centering\includegraphics[width=\linewidth]{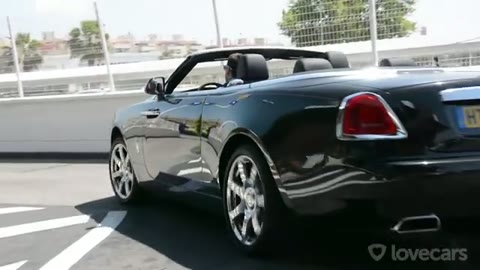}\\{\scriptsize 1.5\,s}\end{minipage}\hfill
\begin{minipage}[t]{0.235\textwidth}\centering\includegraphics[width=\linewidth]{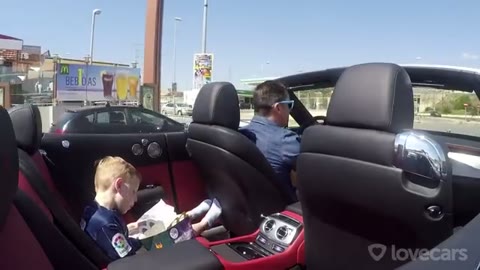}\\{\scriptsize 5.5\,s}\end{minipage}\hfill
\begin{minipage}[t]{0.235\textwidth}\centering\includegraphics[width=\linewidth]{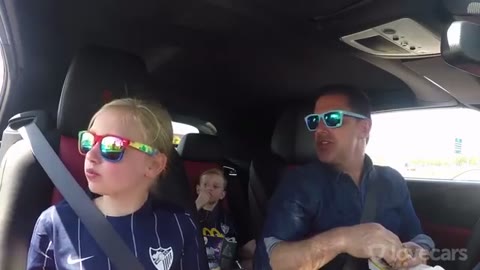}\\{\scriptsize 10.0\,s}\end{minipage}\hfill
\begin{minipage}[t]{0.235\textwidth}\centering\includegraphics[width=\linewidth]{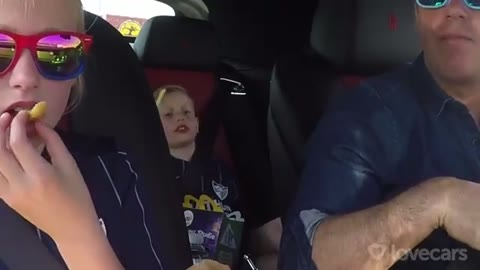}\\{\scriptsize 14.0\,s}\end{minipage}
\\[3pt]
\begin{tabular}{@{}p{1.35cm}@{}p{10.7cm}@{}}
\scriptsize Question & \scriptsize What are the people doing in the car? \\
\scriptsize GT & \scriptsize 0.0--3.7\,s: The man is driving a black convertible car. \; 4.6--7.4\,s: The boy is sitting in the back seat, looking at something and eating something, while the man is driving. \; 8.8--15.7\,s: The man, boy, and girl are all sitting in the car and eating something. \\
\scriptsize Direct & \scriptsize 2\,s: driving a black convertible; \; 6\,s: the boy is sitting and eating; \; 10\,s: the man, boy, and girl are eating; \; 14\,s: \textcolor{pacered}{repeated similar reply} \\
\scriptsize PACE & \scriptsize 2\,s: \textcolor{pacegreen}{emit:} driving a black convertible; \; 6\,s: \textcolor{pacegreen}{emit:} the boy is sitting and eating; \; 10\,s: \textcolor{pacegreen}{emit:} all three are sitting and eating; \; 14\,s: \textcolor{pacegray}{silent: no new information} \\
\end{tabular}
\\[8pt]
\begin{minipage}[t]{0.235\textwidth}\centering\includegraphics[width=\linewidth]{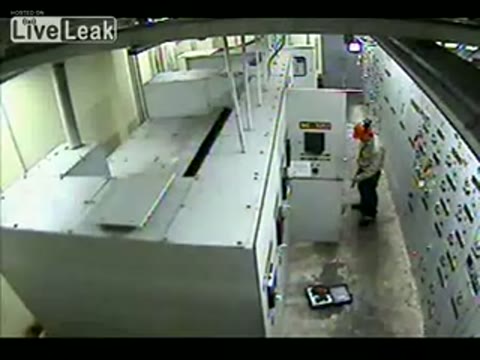}\\{\scriptsize 20\,s}\end{minipage}\hfill
\begin{minipage}[t]{0.235\textwidth}\centering\includegraphics[width=\linewidth]{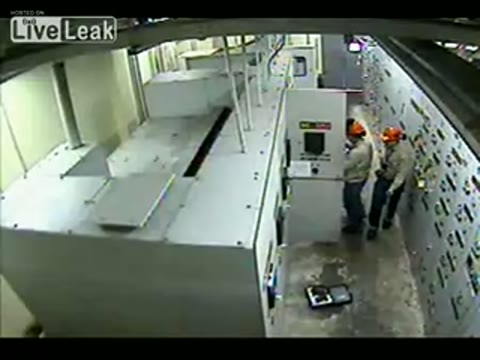}\\{\scriptsize 60\,s}\end{minipage}\hfill
\begin{minipage}[t]{0.235\textwidth}\centering\includegraphics[width=\linewidth]{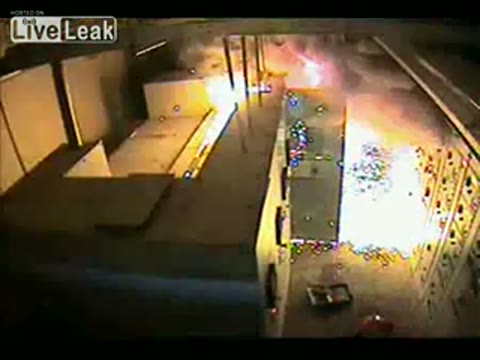}\\{\scriptsize 76\,s}\end{minipage}\hfill
\begin{minipage}[t]{0.235\textwidth}\centering\includegraphics[width=\linewidth]{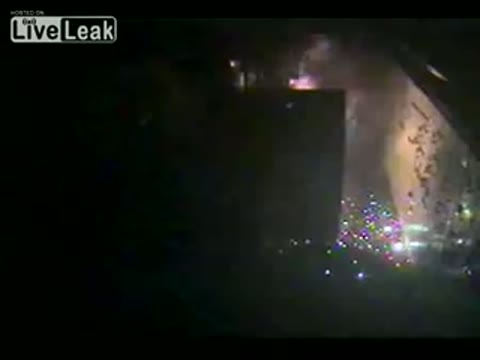}\\{\scriptsize 80\,s}\end{minipage}
\\[3pt]
\begin{tabular}{@{}p{1.35cm}@{}p{10.7cm}@{}}
\scriptsize Question & \scriptsize What suspicious or harmful activities, including unlawful, criminal behaviors or destructive accidents, are happening in the video? \\
\scriptsize GT & \scriptsize 74.3--81.7\,s: The machine exploded, emitting a large amount of sparks and smoke. \\
\scriptsize Direct & \scriptsize 76\,s: \textcolor{pacered}{the machine is exploding and producing sparks}; \; 80\,s: the machine exploded, producing sparks and smoke \\
\scriptsize PACE & \scriptsize 76\,s: \textcolor{pacegray}{hold: the explosion is still unfolding}; \; 80--82\,s: \textcolor{pacegreen}{emit:} the machine exploded, emitting a large amount of sparks and smoke \\
\end{tabular}
\caption{Qualitative cases from ProactiveVideoQA: a rolling web video, top, and a single-truth anomaly clip, bottom. Each case shows sampled frames with their timestamps, the ground-truth replies with their reply windows, the direct chunk-wise replies of the backbone, and the behavior of PACE.}
\label{fig:qualitative}
\end{figure}

Fig.~\ref{fig:qualitative} traces PACE and the direct chunk-wise backbone on two representative videos. On the rolling web video both systems follow the scene as it evolves, and their replies coincide while new content keeps arriving. They part ways once the scene stops changing: at 14\,s the direct protocol produces another near-duplicate of its previous reply, whereas PACE finds that the draft adds nothing beyond its emitted replies and stays silent. On the anomaly clip the divergence is sharper. The direct backbone describes the explosion at 76\,s, while it is still unfolding, so its reply names sparks but misses the smoke that defines the full event. PACE holds the same draft because no completion signal has fired, and emits one complete reply at 80 to 82\,s, after the sparks and smoke have run their course. The two cases exhibit the two behaviors the watch plan distinguishes: rolling updates that track the stream, and a final answer released only once the event has concluded. 

\subsection{Ablations} \label{sec:exp-ablation} Tables~\ref{tab:ablation-gpt} and \ref{tab:ablation-qwen} remove one module at a time: the watch plan, each of the three checks, the plan revision, and the ledger. Ablations are run on ProactiveVideoQA with one closed-source and one open-source backbone, GPT-4.1-mini and Qwen2.5-VL-7B. Because all checks must pass before emission, multiple checks may reject the same compound error, so leave-one-out deltas are a conservative estimate of each module's contribution. We interpret the deltas relative to the pattern predicted by the taxonomy rather than testing individual cells for significance. \begin{table}[!t]
\centering
\small
\caption{Module ablation of PACE on ProactiveVideoQA with the closed-source backbone GPT-4.1-mini. Each row removes one module from the full system; subscripts give the change from the full system.}
\label{tab:ablation-gpt}
\begin{tabular}{@{}lccccc@{}}
\toprule
Variant & WEB & EGO & TV & VAD & Avg. \\
\midrule
PACE full & 54.1\improx & 70.2\improx & 62.8\improx & 56.9\improx & 61.0\improx \\
\;w/o watch plan & $51.4_\detro{2.7}$ & $66.1_\detro{4.1}$ & $59.3_\detro{3.5}$ & $51.9_\detro{5.0}$ & $57.2_\detro{3.8}$ \\
\;w/o stability vote & $52.6_\detro{1.5}$ & $69.0_\detro{1.2}$ & $63.0_\impro{0.2}$ & $55.1_\detro{1.8}$ & $59.9_\detro{1.1}$ \\
\;w/o completion check & $52.3_\detro{1.8}$ & $65.4_\detro{4.8}$ & $59.1_\detro{3.7}$ & $51.6_\detro{5.3}$ & $57.1_\detro{3.9}$ \\
\;w/o grounding check & $53.1_\detro{1.0}$ & $68.5_\detro{1.7}$ & $61.3_\detro{1.5}$ & $54.7_\detro{2.2}$ & $59.4_\detro{1.6}$ \\
\;w/o plan revision & $53.6_\detro{0.5}$ & $69.1_\detro{1.1}$ & $62.0_\detro{0.8}$ & $55.3_\detro{1.6}$ & $60.0_\detro{1.0}$ \\
\;w/o ledger & $51.7_\detro{2.4}$ & $68.2_\detro{2.0}$ & $61.5_\detro{1.3}$ & $54.8_\detro{2.1}$ & $59.1_\detro{1.9}$ \\
\bottomrule
\end{tabular}
\end{table}

\begin{table}[!t]
\centering
\small
\caption{Module ablation of PACE on ProactiveVideoQA with the open-source backbone Qwen2.5-VL-7B. Each row removes one module from the full system; subscripts give the change from the full system.}
\label{tab:ablation-qwen}
\begin{tabular}{@{}lccccc@{}}
\toprule
Variant & WEB & EGO & TV & VAD & Avg. \\
\midrule
PACE full & 59.9\improx & 54.8\improx & 40.1\improx & 33.7\improx & 47.1\improx \\
\;w/o watch plan & $56.4_\detro{3.5}$ & $50.1_\detro{4.7}$ & $36.3_\detro{3.8}$ & $29.2_\detro{4.5}$ & $43.0_\detro{4.1}$ \\
\;w/o stability vote & $57.8_\detro{2.1}$ & $52.8_\detro{2.0}$ & $39.2_\detro{0.9}$ & $31.4_\detro{2.3}$ & $45.3_\detro{1.8}$ \\
\;w/o completion check & $57.7_\detro{2.2}$ & $50.6_\detro{4.2}$ & $36.7_\detro{3.4}$ & $28.9_\detro{4.8}$ & $43.5_\detro{3.6}$ \\
\;w/o grounding check & $57.6_\detro{2.3}$ & $52.2_\detro{2.6}$ & $37.7_\detro{2.4}$ & $30.8_\detro{2.9}$ & $44.6_\detro{2.5}$ \\
\;w/o plan revision & $59.1_\detro{0.8}$ & $53.4_\detro{1.4}$ & $38.9_\detro{1.2}$ & $32.0_\detro{1.7}$ & $45.9_\detro{1.2}$ \\
\;w/o ledger & $56.9_\detro{3.0}$ & $52.4_\detro{2.4}$ & $38.4_\detro{1.7}$ & $31.4_\detro{2.3}$ & $44.8_\detro{2.3}$ \\
\bottomrule
\end{tabular}
\end{table}

\textbf{All modules contribute.} Removing any single module lowers the average PAUC of both backbones, so the overall gain rests on complementary mechanisms rather than a single device. The two largest drops come from the watch plan, 3.8 points on GPT-4.1-mini and 4.1 on Qwen2.5-VL-7B, and the completion check, 3.9 and 3.6 points. Even without the watch plan, the remaining checks keep the system above its unmodified backbone, 57.2 against 55.2 and 43.0 against 41.3, so the plan is necessary for the full gain but not for all of it. 

\textbf{The completion check matters most where events must end.} Its removal costs GPT-4.1-mini 4.8 points on EGO and 5.3 on VAD but only 1.8 on WEB, and Qwen2.5-VL-7B follows the same pattern with 4.2, 4.8, and 2.2 points. WEB questions often admit rolling answers, where a partial reply is already valid output; EGO and VAD hinge on event boundaries, where a model may describe the ongoing action correctly yet cannot fix the final count or outcome before the action ends. This asymmetry supports the central distinction of Sect.~\ref{sec:setup}: support by the current prefix does not make an answer final, and neither stability nor grounding substitutes for an explicit completion judgment.

\textbf{Grounding and stability follow the taxonomy.}
Grounding contributes 1.6 points on GPT-4.1-mini and 2.5 on
Qwen2.5-VL-7B, with the largest per-domain loss on VAD.
Stability contributes 1.1 and 1.8 points; its removal improves
GPT-4.1-mini on TV by only 0.2 points.

\textbf{The ledger and plan revision extend the gains over time.}
Removing the ledger costs most on WEB, 2.4 and 3.0 points,
consistent with its role in cumulative counting and deduplication.
Plan revision contributes 1.0 and 1.2 points, mainly on VAD and EGO,
supporting its role in correcting initial-plan mismatches.

\textbf{Weaker backbones lean on verification, stronger ones on timing.}
Qwen2.5-VL-7B loses more from every removal except the completion
check. This suggests that verification and memory benefit the weaker
backbone more, while completion remains important even for the
stronger model.

\section{Conclusion} \label{sec:conclusion}

%

PACE combines training-free stability, grounding, and completion checks with an auditable event ledger, improving multiple backbones on two proactive benchmarks. Future work should reduce verification cost, diversify verifiers, and learn adaptive watch plans

\section*{Acknowledgments}

This work was supported in part by the Chongqing Natural Science
Foundation (CSTB2024NSCQ-MSX1026), in part by the Fundamental
Research Funds for the Central Universities
(No. SWU-KT\_XJ2025002001), in part by the National Key R\&D
Program of China (No. 2024YFC3808901), and in part by the
NSFC (62102328).

\bibliographystyle{splncs04}
\bibliography{refs}

\end{document}